\documentclass{article}

\usepackage{microtype}
\usepackage{graphicx}
\usepackage{subfigure}
\usepackage{booktabs} % for professional tables

\usepackage{amsmath}
\usepackage{makecell}
\usepackage{multirow}
\usepackage{enumitem}% http://ctan.org/pkg/enumitem

\usepackage{colortbl}
\usepackage[dvipsnames]{xcolor}
\definecolor{green_im}{rgb}{0.1, 0.55, 0.3}
\newcommand{\drop}[1]{\textcolor{red}{\tiny{$\downarrow$#1}}}
\newcommand{\rise}[1]{\textcolor{green_im}{\tiny{$\uparrow$#1}}}

\definecolor{Gray}{gray}{0.85}
\definecolor{LightCyan}{rgb}{0.88,1,1}
\newcolumntype{a}{>{\columncolor{Gray}}c}
\newcolumntype{b}{>{\columncolor{white}}c}

\DeclareMathOperator*{\argmin}{arg\,min}

\usepackage{hyperref}
\usepackage{subfigure}
\usepackage{parskip}

\usepackage[algo2e]{algorithm2e}

\usepackage[accepted]{icml2023}

\usepackage{amsmath}
\usepackage{amssymb}
\usepackage{mathtools}
\usepackage{amsthm}

\usepackage[capitalize,noabbrev]{cleveref}

\theoremstyle{plain}
\newtheorem{theorem}{Theorem}[section]

\newtheorem{lemma}[theorem]{Lemma}

\theoremstyle{definition}

\theoremstyle{remark}

\usepackage[textsize=tiny]{todonotes}

\usepackage{xspace}
\newcommand*{\eg}{\emph{e.g.}\@\xspace}

\newcommand*{\ie}{\emph{i.e.}\@\xspace}

\newcommand*{\vs}{\emph{v.s.}\@\xspace}

 \newcommand{\system}{Open-VCLIP\xspace}

\icmltitlerunning{Transforming CLIP to an Open-vocabulary Video Model}

\begin{document}

\twocolumn[
\icmltitle{Open-VCLIP: Transforming CLIP to an Open-vocabulary Video Model \\ via Interpolated Weight Optimization}



\icmlsetsymbol{equal}{*}

\begin{icmlauthorlist}
\icmlauthor{Zejia Weng}{aaa,bbb}
\icmlauthor{Xitong Yang}{ccc}
\icmlauthor{Ang Li}{ddd}
\icmlauthor{Zuxuan Wu}{aaa,bbb}
\icmlauthor{Yu-Gang Jiang}{aaa,bbb}

\end{icmlauthorlist}

\icmlaffiliation{aaa}{Shanghai Key Laboratory of Intelligent Information Processing, School of Computer Science, Fudan University}
\icmlaffiliation{bbb}{Shanghai Collaborative Innovation Center of Intelligent Visual Computing}
\icmlaffiliation{ccc}{Meta AI}
\icmlaffiliation{ddd}{angli.ai}

\icmlcorrespondingauthor{Zuxuan Wu}{zxwu@fudan.edu.cn}

\icmlkeywords{Machine Learning, ICML}

\vskip 0.3in
]



\printAffiliationsAndNotice{}  % leave blank if no need to mention equal contribution

\begin{abstract}
Contrastive Language-Image Pretraining (CLIP) has demonstrated impressive zero-shot learning abilities for image understanding, yet limited effort has been made to investigate CLIP for zero-shot video recognition. We introduce \system, a simple yet effective approach that transforms CLIP into a strong zero-shot video classifier that can recognize unseen actions and events at test time. 
Our framework extends CLIP with minimal modifications to model spatial-temporal relationships in videos, making it a specialized video classifier, while striving for generalization.  
We formally show that training an Open-VCLIP is equivalent to continual learning with zero historical data. To address this problem, we propose \textit{Interpolated Weight Optimization}, which utilizes the benefit of weight interpolation in both training and test time. We evaluate our method on three popular and challenging action recognition datasets following various zero-shot evaluation protocols and we demonstrate our approach outperforms state-of-the-art methods by clear margins. In particular, we achieve 87.9\%, 58.3\%, 81.1\% zero-shot accuracy on UCF, HMDB and Kinetics-600 respectively, outperforming state-of-the-art methods by 8.3\%, 7.8\% and 12.2\%. Code is released at {\small\url{https://github.com/wengzejia1/Open-VCLIP}}.

\end{abstract}

\section{Introduction}

Zero-shot learning is a challenging problem that requires deep neural networks to recognize novel unseen classes during testing without having seen them during training. The generalization ability of classifying new classes without the need for manual annotations makes it particularly useful in real-world applications.
% where images and labels are extremely difficult to obtain. 
While extensive studies have been conducted on zero-shot learning~\cite{zellers2017zero,brattoli2020rethinking,xu2017transductive}, CLIP~\cite{radford2021learning} recently demonstrates surprising zero-shot abilities in a variety of tasks, such as image segmentation~\cite{wang2022cris,ghiasi2021open,ghiasi2022scaling}, image editing~\cite{zheng2022bridging,crowson2022vqgan}, by pretraining on web-scale image and text pairs in a contrastive manner. 

\begin{figure}[t]
    \centering
    \includegraphics[width=0.5\textwidth]{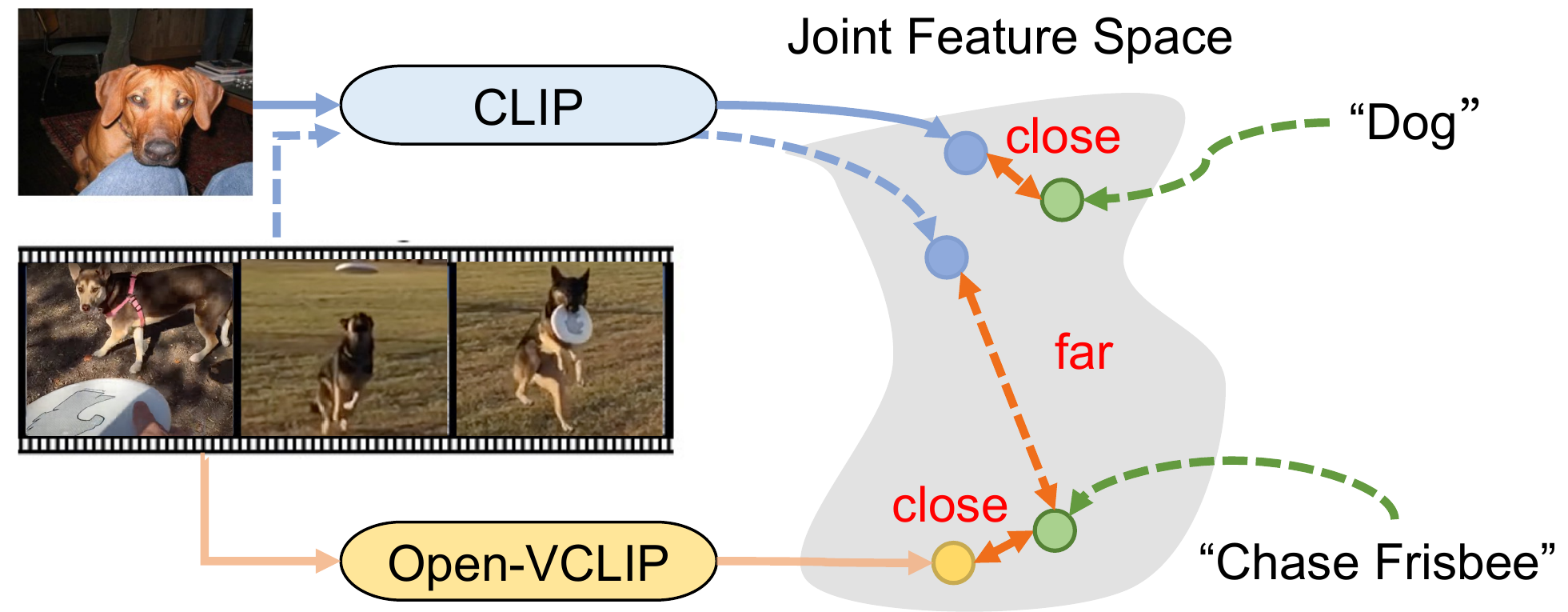}
    \caption{While CLIP has shown impressive results for zero-shot image recognition, it cannot effectively recognize novel actions in videos. This paper aims to transform CLIP to a strong zero-shot video classifier with minimal modifications. }
    \label{fig:pipeline}
\end{figure}

While significant zero-shot results are achieved in the image domain, limited effort has been made to explore CLIP for zero-shot video action recognition. Extending CLIP, designed for image tasks, to the video domain, is extremely challenging, particularly in the zero-shot setting.  On one hand, to better recognize actions and events in videos, the rich temporal dynamics encoded in videos are expected to be carefully captured. Although one could treat videos as a bag of frames and perform temporal pooling over frame-based predictions~\cite{wang2021actionclip},  it is found that fine-tuning pretrained models like CLIP with specialized temporal modeling components on top of off-the-shelf image models produces better results~\cite{ni2022expanding}. However, the improved results come at the cost of reduced generalization as optimizing specialized parameters will inevitably requires fine-tuning the pretrained CLIP model.  The derived model then tends to overfit to the video dataset used for fine-tuning, which are substantially smaller than the image-text dataset used to train CLIP.  As a result, the zero-shot ability of CLIP diminishes over the course of fine-tuning. 

It is worth noting that the start point of the fine-tuning process is CLIP, an image classifier with strong zero-shot abilities, while the end point is a specialized model for video understanding. This motivates us to seek a middle ground between generalization and specialization---adapting a pretrained CLIP model to the video domain, and the resulting model is expected to not only recognize known actions and events seen during training but also enjoy the zero-shot learning abilities as CLIP for novel video categories. Interestingly, we discover from a theoretical perspective that this problem is essentially a continual learning problem, which aims to adapt a pretrained model to new tasks with less forgetting on previous knowledge. Traditional continual learning typically seeks decent performance on all seen tasks with access to historical data~\cite{hu2021opin,balaji2020taskonomycl,shin2017continual}. However, this is particularly challenging for adapting CLIP as raw data used for training are private. Furthermore, our goal is slightly beyond continual learning: we hope to adapt CLIP to be a strong zero-shot video learner that generalizes well to unknown video actions and events, rather than exactly preserving its knowledge for image tasks, which again is difficult without access to the training data of CLIP.

With this in mind, we explore the problem of constructing an open-vocabulary video CLIP by simply leveraging the pretrained weights that are publicly available. More specifically, we build upon CLIP with minimal modifications so as to derive a video action recognition model that can not only capture temporal information among different frames but also generalizes well to unseen actions and events during testing. To optimize the continual learning-based training formulation, we propose a novel method called \textit{Interpolated Weight Optimization} which regularizes the fine-tuning process of the CLIP model by adding a link between the original CLIP model and the current model. This prevents the derived model from drifting away from CLIP, which we argue is beneficial for generalization. Furthermore, in addition to training, we also link derived optima along the optimization trajectory at test time for improved generalization.

We conduct extensive experiments to evaluate the performance of \system. In particular, we use Kinetics-400 as a proxy dataset to extend CLIP to the video domain and evaluate its zero-shot performance on UCF-101, HMDB-51 and Kinetics-600 with various evaluation protocols. \system achieves state-of-the-art zero-shot video action recognition performance, offering 87.9\%, 58.3\%, 81.1\% zero-shot accuracy on UCF, HMDB and Kinetics-600 respectively, which are 8.3\%, 7.8\% and 12.2\% better compared to alternative methods. Furthermore, \system also achieves the best trade-off between close-set and zero-shot performance across all the benchmarks.

\section{Related Work}
\textbf{Zero-shot Video Action Recognition.} 
Zero-shot video action recognition requires models to recognize new actions that are not seen during training, which is useful in real-world applications where data and their corresponding labels are difficult to collect.
Early work focuses more on how to represent actions properly. For example, manually-crafted attributes \cite{liu2011recognizing,zellers2017zero}, object features   \cite{jain2015objects2action,gao2019know} are used to represent actions. Researchers  also use word embeddings of actions \cite{brattoli2020rethinking,xu2017transductive} as textual semantic representations.
Recently, pretraining with large-scale vision-text data is gaining attention, as it achieves impressive results for zero-shot image classification \cite{radford2021learning,jia2021scaling}. There is also a plethora of work using knowledge learned in large-scale pretrained vision-language models to down-stream tasks in a zero-shot manner~\cite{wang2022cris,ghiasi2021open,ghiasi2022scaling}. While extensive studies have been conducted for zero-shot image understanding, zero-shot video classification remains less explored. ActionCLIP~\cite{wang2021actionclip} and XCLIP experiment with zero-shot setting for action recognition~\cite{ni2022expanding}, but use the same strategy as in images for zero-shot learning and ignore the forgetting problem when adapting the CLIP model. In contrast, our goal is to explicitly build a strong open-vocabulary zero-shot video classifier by regularizing the fine-tuning process.

\textbf{Continual Learning.}
Continual learning aims at training a model on multiple sequential tasks without catastrophically forgetting knowledge from previous tasks. 
Existing approaches can be divided into three categories: memory-based, expansion based and regularization-based methods. Memory-based methods typically utilize a replay buffer to store past examples or related information such as gradients \cite{farajtabar2019orthogonal}. In addition, memory replay is effective for continual learning \cite{hu2021opin,balaji2020taskonomycl}, however, in our case, it is not feasible to directly use these approaches as historical data are not available. Expansion-based methods \cite{pnn} expand the network over time in order to maintain past performance. Closest to our approach, perhaps are regularization-based methods \cite{yin2020sola} such as EWC \cite{ewc}, which add a regularization term to the optimization that constrains new model weights to be close to the original ones. But still historical data need to be used to calculate the Fisher information matrix for approaches like~\cite{ewc}. 
Our approach differs from standard continual learning in that we wish to transfer the knowledge, \ie the ability to perform zero-shot learning, from a previous image task to video tasks without access to historical data at all.

\section{Preliminary: Video Action Recognition Using CLIP}
\label{sec:pre}
Video action recognition is a fundamental yet challenging task that often requires intensive model training on large-scale datasets. Inspired by the success of contrastive language-image pretraining (CLIP)~\cite{radford2021learning}, recent work has proposed to fine-tune the well-trained CLIP model on the target video dataset and has achieved state-of-the-art results~\cite{xu2021videoclip,wu2022transferring}.

To adapt CLIP for video action recognition (VCLIP), a common strategy~\cite{arnab2021vivit,bulat2021space,bertasius2021space,xing2022svformer} is to extend the image encoder to capture temporal dynamics in videos and, in the meanwhile, align the video representation with the text representations of its corresponding label (\eg, ``playing drums"). Specifically, given a video clip $V\in \mathcal{V}_B$ and an action label described in textual prompts $T\in \mathcal{T}_B$, the goal of fine-tuning is to maximize the similarity:
\begin{align}
    \text{sim}(v, t) = \frac{\langle v, t\rangle}{\|v\| \|t\|}, \;\;\; v=f_{\theta_B^V}(V), \; t=f_{\theta_B^T}(T),
\end{align}
% if $V$ and $T$ represent the same video. As the text encoder is typically frozen ($\theta_B^T = \theta_A^T$)~\cite{ilharco2022patching,thengane2022clip}, the fine-tuning stage primarily focuses on optimizing the image encoder $\theta_B^V$ for adaptation to the video domain. We drop the superscript in the subsequent paragraphs for brevity.
if $V$ and $T$ represent the same video. Here, $\mathcal{V}_B$ denotes the video dataset for task $B$, $\mathcal{T}_B$ denotes the corresponding label set, $f_{\theta_B^T}$ denotes the text encoder and $f_{\theta_B^V}$ denotes the visual encoder. As the text encoder is typically frozen ~\cite{ilharco2022patching,thengane2022clip}, the fine-tuning stage primarily focuses on optimizing the visual encoder for adaptation to the video domain. We drop the superscript in the subsequent paragraphs for brevity.

\section{Our Approach}
We introduce our approach in this section which consists of two major components: (1) constructing a VCLIP model from the image-based CLIP in order to better explore spatial-temporal relationships in videos, and (2) regularizing the fine-tuning process so that the derived model can generalize well to unseen actions and events.

\subsection{Constructing VCLIP for Video Understanding}\label{al:temporalattend}
Temporal relations among video frames contain important information for identifying actions. The image CLIP model does not have the ability to aggregate temporal features and thus is no longer suitable for video tasks. So we devote to injecting temporal modeling ability to the original CLIP model for a better transfer from images to videos.

Adding additional temporal modeling networks is the most convenient way to achieve this, however, this generally incurs extra parameters that are computationally expensive and makes it harder for weight interpolation that greatly benefits zero-shot abilities, as will be described later.
 We observe that the self-attention layer in vision transformer is quite scalable,  operating on image frame patches as:
\begin{align}
    y_{s,t} = \text{Softmax}(\frac{q_{s,t} K_t^\text{T}}{\sqrt{d}}) V_t,
\end{align}
where $d$ refers to the dimention of vectors, $q_{s,t}$ refers to the query vector of the $s$-th token in the $t$-th frame, $K_t^\text{T}$ is the transpose of the matrix composed of key vectors in the $t$-th frame, and $V_{t}$ is the matrix composed of value vectors in the $t$-th frame. Obviously, each token will only obtain information from its belonging frame. To overcome this issue, we expand the temporal attention view for every self-attention layer for aggregating the global temporal information thanks to the stacking of self-attention layers. The new self-attention layer is implemented as follows:
\begin{align}
    y_{s,t} = \text{Softmax}\left(\frac{q_{s,t} [K_{(t-1)\sim(t+1)}]^\text{T}}{\sqrt{d}}\right) [V_{(t-1)\sim(t+1)}].
\end{align}
At this time, each patch will obtain information from its belonging frame and its neighbouring frames. The special modification lies in $[K_{(t-1)\sim(t+1)}]$ and $[V_{(t-1)\sim(t+1)}]$ which refers to the Key/Value matrix be composed with key/value vectors belonging to not only the $t$-th frame, but also the neighbour frames. Such a small modification helps the model to gain better temporal information aggregation ability, while perfectly fitting our algorithm since no extra parameters are added.

\subsection{Training Open-VCLIP}
 We now introduce the training method with a carefully designed optimization strategy for improved zero-shot video classification. We start by formulating the problem from its original purpose and then derive its approximation which leads to a challenging continual learning problem with zero historical data. We propose a novel regularization-based method to optimize VCLIP, named \textit{Interpolated Weight Regularization} (IWR). A model-averaging approach called Stochastic Weight Averaging (SWA) is further incorporated to improve model generalizability.

\subsubsection{Problem Definition}
While fine-tuning the CLIP model yields impressive results for close-set video classification~\cite{wang2021actionclip,ni2022expanding}, its performance on unseen categories is poor---might be worse than the original CLIP model as shown in \cref{table:baseline-compare}. Below, we elaborate on how to construct a robust \textit{open-vocabulary} video model from a pretrained CLIP model.

Formally, our goal is to obtain the optimal vision encoder $f_{\theta_U}$ that satisfies:
\begin{align}
\theta_U = \arg\min_{\theta} L(\theta;D_U).
\label{eq:universal}
\end{align}
Here, $D_U = \left\{\mathcal{V}_U, \mathcal{T}_U\right\}$ is a \emph{hypothetical} universal dataset that contains all possible videos and their corresponding text descriptions (\ie, action labels). $L$ is the loss function defined on the video-text pairs.
Optimizing such an objective directly is infeasible, yet luckily it can be approximated by training on a large-scale dataset with sufficiently diverse video and text data. 

\subsubsection{An Approximation Equivalent to Continual Learning}
We first consider a video action recognition dataset $D_B = \left\{\mathcal{V}_B, \mathcal{T}_B \right\}$ used for fine-tuning. Unfortunately, even though the abundant video data $\mathcal{V}_B$ serve as a good approximation of $\mathcal{V}_U$, its text space is extremely limited, bounded by the number of annotated action categories (\eg, $|\mathcal{T}_B|=400$ for Kinetics-400~\cite{kay2017kinetics}). As a result, the video representation is prone to overfitting to the highly skewed text space and the zero-shot ability of the CLIP model diminishes over the fine-tuning process. On the other hand, the scale of the image training dataset for CLIP, is sufficiently large to approximate $D_U$, but there exists a domain gap between the image space and the video space.
Even worse, this dataset is a private dataset and only the CLIP model weights $\theta_A$ are accessible for fine-tuning.

With these pros and cons in mind, we now seek to leverage \textit{both} $\theta_A$ and $D_B$ to construct a strong open-vocabulary model. We believe that $\theta_A$ should contain useful information from the large image dataset that CLIP was originally trained on.
Following this intuition, we notice that the initial VCLIP model (without any fine-tuning) $\theta_A$ is indeed an optimal solution to a large-scale video dataset $D_{A}$ with a sufficiently diverse text space $\mathcal{T}_{A}$ (\cref{lemma1}).

\begin{lemma} Suppose the image CLIP model was trained on an image-text dataset $D_{\bar{A}} =\{\mathcal I_{\bar{A}},\mathcal T_{\bar{A}}\}$ with $N$ examples. Then, there exists a diverse video-text dataset $D_A=\{\mathcal V_A, \mathcal T_A\}$ containing $N$ examples where the video CLIP model with original CLIP parameters $\theta_A$ is optimal. 
\begin{proof}See Appendix.
\end{proof}\label{lemma1}
\end{lemma}

Although $D_A$ is unknown, it is helpful to understand our problem. Given the fact that $\theta_A$ is an optimal solution in a large scale dataset $D_A$, a natural idea for approximating the universal objective is to combine both datasets in the derivation, although $D_A$ is unknown in reality.

Following this idea, \cref{eq:universal} is transformed into:
\begin{align}
\arg\min_{\theta} L(\theta;D_A) + L(\theta;D_B) 
\label{eq:estimation}
\end{align}
where $D_A$ is completely unknown and only $D_B$ is present. However, we have the optimal solution $\theta_A$ on $D_A$. In that case, the formulation becomes equivalent to \emph{continual learning}, \ie, continually training the model on a new dataset while preserving its performance on historical data so as to achieve sufficient generalizability.

\subsubsection{Interpolated Weight Regularization}
While there have been many methods of studying continual learning, most of them are based on storing historical data or information, which is not applicable in our case. It is due to the fact that we do not have access to $D_A$ at all.
Inspired by empirical results in~\cite{ilharco2022patching}, we propose to lowing the first loss term in \cref{eq:estimation} by introducing an \textit{optimization-free} weight interpolation strategy:
\begin{equation}
\begin{aligned}
&\theta=\lambda \theta_A + (1-\lambda)\theta_B,
\label{eq:patch}
\end{aligned}
\end{equation}
where $\lambda$ is a trade-off hyperparameter. \cref{eq:patch} is called model patching~\cite{ilharco2022patching}, which is commonly used between two converged model weights, \ie, $\theta_B$ is trained separately on $D_B$ only. A potential risk of this method is that we have no explicit optimization on the curve fitting performance of the final patched model, \ie, the patch may be severely underfitting to $D_B$ or sensitive to the trade-off parameter $\lambda$.

To address this issue, we propose Interpolated Weight Regularization as part of the training procedure, which regularizes the loss of the interpolated weights on $D_B$. Given that the $\lambda$ is a hyperparameter that may vary, instead of optimizing a single point, we look for a solution such that the patched model's performance \emph{w.r.t.} a range of interpolation coefficients are optimized. This is achieved by sampling a balancing coefficient during training. The final optimization objective becomes:
\begin{equation}
    \argmin_{\theta_B} \ \mathcal L = L(\theta_B; D_B) + \beta {L(\alpha \theta_A + (1-\alpha) \theta_B; D_B)}
    \label{eq:final_linear_connect}
\end{equation}
where $\alpha \sim \texttt{U}(0, \lambda)$. The $(0,\lambda)$ interval corresponds to the low-level region between the interpolated and end weights. $\beta$ is the trade-off coefficient for regularization and it is set as $\beta=C\frac{1}{1-\alpha}$ in practice where $C$ is a constant value. The loss can be optimized by calculating its derivative as follows:
\begin{align}
    \frac{d\mathcal L}{d\theta} &= \left.\frac{dL}{d\theta}\right|_{\theta=\theta_B} + \beta(1-\alpha)\left.\frac{dL}{d\theta}\right|_{\theta=\alpha\theta_A+(1-\alpha)\theta_B}\\
    &= \left.\frac{dL}{d\theta}\right|_{\theta=\theta_B} + C\left.\frac{dL}{d\theta}\right|_{\theta=\alpha\theta_A+(1-\alpha)\theta_B}~.
    \label{eq:final}
\end{align}
After we obtain the optimal $\theta_B$, \cref{eq:patch} is applied to achieve the final VCLIP model weights.

\subsubsection{Stochastic Weight Averaging}

Improved zero-shot predicting ability lies in good generalization ability. We further introduce an upgrade to the above method by applying Stochastic Weight Average (SWA) on the interpolated weights along the training process to find the ``flat minimum'', which refer to a region in the weight space where the loss function is relatively flat and the test error is relatively low according to \cite{izmailov2018averaging}. \cref{eq:swa} showcases our solution:
\begin{align}
    \sum_i^N \frac{{\lambda\theta_A+(1-\lambda)\theta_i}}{N} = \lambda \theta_A + (1-\lambda)\, \boxed{\frac{1}{N} \sum_i^N \theta_i}
    \label{eq:swa}
\end{align}
where $\theta_i$ refers to the $i$-th set of parameters we select during the training process. As \cref{eq:swa} shows, the moving average of the interpolated weights equals to interpolating the moving average of the updated weights, showing the order of the SWA and the weight interpolation is interchangable. So, in practice, we maintain the moving average of the model's weights during training and do weight interpolation in the end. By averaging model weights from the optimization trajectories, we will get a robust model with better generalization to unseen classes.

\subsection{Algorithm Summary}

We show implementation of \system in the following pseudo code.

\begin{algorithm}[h]\small
   \caption{Training}
   \label{alg:example_training}
\begin{algorithmic}
   \STATE {\bfseries Input:} Dataset $D=\{V_i, y_i\}^{N}$, Model $f_{\theta}$, Step = 0. 
   \STATE {\bfseries Require:} SWA begins at $T$ step with a cycle length $c$. SWA Param ${\theta_{\texttt{SWA}}}$. Counting flag $l=0$. Model Param $\theta$ is initialized by the CLIP Param $\theta_\texttt{CLIP}$. Hyper-param R, $\beta$. Learning rate $\delta$.\\
   \REPEAT
   \STATE Step $\leftarrow$ Step + 1 \\
   \STATE sample  $\{V_i, y_i\}^M \subseteq D$, $\widetilde{V} \leftarrow \{V_i \}^M$, $\widetilde{y} \leftarrow \{y_i \}^M$ \\
   \tcp{normal supervision loss.}
   \STATE $L(f_{\theta}) \leftarrow  \frac{1}{M} \sum_{i=1}^M L(\widetilde{V}_i,\widetilde{y}_i, \theta)$\\ 
   \tcp{interpolation regularization.}
   \STATE Sample $\alpha \sim \texttt{Uniform(0, R)}$
   \STATE Initialize $\widetilde{\theta} \leftarrow  \alpha \cdot \theta_\texttt{CLIP} + (1-\alpha) \cdot \theta$ 
   \STATE $L(f_{\widetilde{\theta}}) \leftarrow  \frac{1}{M} \sum_{i=1}^{M} L(\widetilde{V}_i,\widetilde{y}_i, \widetilde{\theta})$ \\
   \tcp{update model with combined loss.}
   % \STATE $L_{total} = L(f_{\theta}) + \beta \cdot L(f_{\widetilde{\theta}})$
   $\theta \leftarrow  \theta - \delta \nabla_{\theta} (L(f_{\theta}) + \beta \cdot L(f_{\widetilde{\theta}})) $ \\
   \quad \\
   \tcp{stochastic weight average.}
   \IF{$\text{Step} > T$ and $\text{mod}(\text{Step} - T, c) == 0$}
   \STATE ${\theta_{\texttt{SWA}}} \leftarrow ({\theta_{\texttt{SWA}}}*l+\theta) / ({l+1})$
   \STATE $l \leftarrow l + 1$
   \ENDIF
   \UNTIL{converge}
\end{algorithmic}
\end{algorithm}

\begin{algorithm}[h]\small
   \caption{Inference}
   \label{alg:example_testing}
\begin{algorithmic}
   \STATE {\bfseries Input:} Testing Set $D=\{V_i\}^{N}$, Model Param ${\theta_{\texttt{SWA}}}$, Original CLIP Param ${\theta_{\texttt{CLIP}}}$.
   \STATE {\bfseries Require:} Set Interpolation Ratio as $\lambda$\\
   \STATE {\bfseries Initialize:} $\theta_{\texttt{FINAL}} = \lambda * \theta_\texttt{CLIP} + (1-\lambda) * {\theta_{\texttt{SWA}}}$ \\
   \STATE $\text{Predict} = f(D; \theta_{\texttt{FINAL}})$
   % \REPEAT
   % \STATE sample  $\{V_i\}^M \subseteq D$, $\widetilde{V} \leftarrow \{V_i \}^M$ \\
   % $\text{Predict} = f(\widetilde{V}; \theta_{\texttt{FINAL}})$
   % \UNTIL{go through all test data}
\end{algorithmic}
\end{algorithm}

\begin{table*}[t!]
\caption{Zero-shot classification performance for various algorithms on UCF and HMDB with different protocols (see \cref{evaluation_protocol} ).}

\label{table:sota-compare}
\vskip 0.15in
\begin{center}
\begin{small}
\begin{sc}
\begin{tabular}{lccccc}
\toprule
\multirow{2}{*}{Method}  & \multirow{2}{*}{encoder} & \multicolumn{2}{c}{UCF} & \multicolumn{2}{c}{HMDB} \\
\cmidrule(lr){3-4}\cmidrule(lr){5-6}
& & EP1 & EP2 & EP1 & EP2 \\ 
\midrule
GA \cite{mishra2018generative} & C3D & 17.3$\pm$1.1 & - & 19.3$\pm$2.1 & -\\
TARN \cite{bishay2019tarn} & C3D & 19.0$\pm$2.3 & - & 19.5$\pm$4.2 & - \\
CWEGAN \cite{mandal2019out} & I3D & 26.9$\pm$2.8 & - & 30.2$\pm$2.7 & -  \\
TS-GCN \cite{gao2019know} & GLNet  & 34.2$\pm$3.1 & - & 23.2$\pm$3.0 & -  \\
PS-GNN \cite{gao2020learning} & GLNet & 36.1$\pm$4.8 & - & 25.9$\pm$4.1 & - \\
E2E \cite{brattoli2020rethinking}  & R(2+1)D  & 48.0 & 37.6  & 32.7 / 26.9 \\
DASZL \cite{kim2021daszl} & TSM & 48.9$\pm$5.8 & - &  - & - \\
ER \cite{chen2021elaborative} & TSM & 51.8$\pm$2.9 & - & 35.3$\pm$4.6 & - \\
ResT \cite{lin2022cross} &ResNet101 & 58.7$\pm$3.3 & 40.6 & 41.1$\pm$3.7 & 34.4 \\
\midrule 
ActionCLIP \cite{wang2021actionclip}  & ViT-B/16&  - & 69.5 &  - & 50.5\\
Text4Vis \cite{wu2022transferring} & ViT-L/14 & 85.8$\pm$3.3 & 79.6 &  58.1$\pm$5.7 & 49.8 \\
\midrule 
\makecell{\multirow{3}{*}{\system}} & ViT-B/32 & 87.1$\pm$2.4 & 79.5 &  62.3$\pm$4.0 & 49.9  \\
 & ViT-B/16 &  89.9$\pm$1.7 & 83.5 &  64.5$\pm$4.5 & 53.2 \\
 & ViT-L/14 &  93.1$\pm$1.9 & 87.9 &  68.5$\pm$4.0 & 58.3 \\
\bottomrule
\end{tabular}
\end{sc}
\end{small}
\end{center}
\vskip -0.15in
\end{table*}

\section{Experiments}
We present in this section our experimental evaluation. We compare our Open-VCLIP method with state-of-the-art video models and perform ablation studies to reveal the characteristics of our method.

\subsection{Experimental Setup}
\subsubsection{Datasets}

In our experiments, we use the following four datasets:

\textbf{Kinetics-400\&600:} Kinetics-400 \cite{kay2017kinetics} and Kinetics-600 \cite{carreira2018short} datasets are large-scale datasets for human action recognition, containing 400 and 600 action classes, respectively. The Kinetics-600 dataset is an extension of the Kinetics-400 dataset, including 220 new categories. These additional classes provide a valuable resource for evaluating the zero-shot ability of models trained on the Kinetics-400. Each video is a 10-second around action moment annotated from YouTube videos. 
 
\textbf{UCF-101:} The UCF-101 dataset \cite{soomro2012ucf101} is a widely used dataset for human action recognition, which contains 13,320 video clips from 101 action categories. Each video in the dataset is a short clip of an action, with an average length of 7.21 seconds, that is captured from real-world scenarios. Officially, three different training/testing splits files are provided.

\textbf{HMDB-51:} HMDB-51 \cite{kuehne2011hmdb}  contains 6,849 clips divided into 51 action categories with each category containing at least 101 clips. Officially, three different training/testing splits are provided to evaluate the model. To ensure a balance in the number of samples per category, 1,746 videos are left ``unused'' in each split, while each category is guaranteed to have 70 training samples and 30 test samples. Having a consistent number of testing samples for each category allows for a fair evaluation.

We use the Kinetics-400 dataset as the training set and use UCF-101, HMDB-51, and a subset of the Kinetics-600 dataset as test datasets. By testing on datasets that contain no/few overlapping categories with the training dataset, we are able to provide a realistic and comprehensive evaluation of the zero-shot capabilities. Detailed evaluation protocols are described below.

\subsubsection{Evaluation protocols} \label{evaluation_protocol}

\textbf{UCF-101\&HMDB-51:} Following \cite{brattoli2020rethinking} and \cite{ni2022expanding}, there are three protocols for zero-shot evaluation on these two datasets.  
\begin{itemize}[itemsep=0pt,topsep=0pt]
    \item Evaluation Protocol 1 (\textit{EP1}): We first randomly choose half of the classes in the test dataset,  \ie, 50 for UCF and 25 for HMDB, and repeat ten times and report averaged results for each dataset. \cite{brattoli2020rethinking}
    \item Evaluation Protocol 2 (\textit{EP2}): We test the model on full UCF and HMDB, \ie, evaluating on all 101 UCF classes and all 51 HMDB classes \cite{brattoli2020rethinking}. 
    \item Evaluation Protocol 3 (\textit{EP3}): We perform testing using three official splits and average the results of each split \cite{ni2022expanding}. The average top-1 accuracy and standard deviation are reported.
\end{itemize}

\textbf{Kinetics-600:} \cite{chen2021elaborative} randomly split 220 new classes in Kinetics-600 into 60 validation classes and 160 testing classes, respectively, three times. We use the same splits provided by \cite{chen2021elaborative}. We report the average top-1 accuracy as well as the standard deviation.

\subsubsection{Implementation Details}

The initial learning rate is set to $3.33\times10^{-6}$ and is decayed to $3.33\times10^{-8}$ following the cosine decay scheduler. We use 2 epochs to warm up the training and another 20 epochs for fine-tuning on Kinetics-400. Th learning rate for warm-up is set to $3.33\times10^{-8}$ and we use 8 GPUs for training, each contains a batch of 8 samples. Since we are performing vision-text alignment during the training process, augmentations such as mixup and cutmix significantly impact the alignment process. Therefore, we do not use them to avoid any potential negative effects for alignment. Instead, we simply use basic augmentations like color jittering, random flipping, and random cropping. 
Each video clip is extracted every 16 frames for a total of 8 frames to form an input video clip. During testing, 3 clips with 1 crop (``$3\times1$ views'') per video will be sampled to produce a prediction and the results will be further aggregated linearly. Furthermore, we set the interval of regularization to (0.0, 0.6), the balance ratio $C$ of the regularization is set to 0.5, and we start SWA from the second epoch when the warm-up stage finishes.

\subsection{Main Results}

\subsubsection{Comparison to state-of-the-arts}
We compare our method to state-of-the-art zero-shot video action recognition methods using the UCF-101, HMDB-51, and a subset of Kinetics-600 datasets for evaluation. We first present the results under \textit{Evaluation Protocol 1} and \textit{Evaluation Protocol 2}, summarized in \cref{table:sota-compare}. The first block of \cref{table:sota-compare} presents methods that do not rely on large-scale vision-language pretrained models, while the last two blocks of \cref{table:sota-compare} show the performance of CLIP-based methods 
that transfer the image CLIP model to the video domain by fine-tuning on the Kinetics-400 dataset. 

As shown in the table, the results of CLIP-based methods, \ie ActionCLIP, TEXT3VIS, and ours, are significantly better than other methods. This indicates that knowledge from large-scale vision-language pretrained models is important for zero-shot video action recognition performance. Furthermore, when using the ViT-B/16 encoder, our method outperforms ActionCLIP by 14\% (83.5\% vs 69.5\%) and 2.7\% (53.2\% vs 50.5\%); when using the ViT-L/14 encoder, our method outperforms TEXT4VIS by 8.3\% (87.9\% vs 79.6\%) and 8.5\% (58.3\% vs 49.8\%). These substantial improvements in accuracy highlight that our approach can effectively adapt the CLIP model to the video domain for improved zero-shot recognition.
We also compare with X-CLIP \cite{ni2022expanding} under \textit{Evaluation Protocol 3} on UCF and HMDB datasets. Results in \cref{table:protocol3} reveal that \system outperforms X-CLIP by 11.4\% (83.4\% vs 72.0\%) on UCF and 9.3\% (53.9\% vs 44.6\%) on HMDB using ViT-B/16, suggesting the effectiveness of our approach. 

\begin{table}[h!]
\caption{We compare the zero-shot classification performance with X-CLIP on UCF and HMDB dataset with Evaluation Protocol 3.}
\label{table:protocol3}
\vskip 0.15in
\begin{center}
\begin{small}
\begin{sc}
\begin{tabular}{lcccr}
\toprule
Method  & encoder & UCF & HMDB \\
\midrule
X-CLIP  & ViT-B/16& 72.0$\pm$2.3 & 44.6$\pm$5.2\\
\midrule
\makecell{\multirow{3}{*}{\system}} & ViT-B/32 & 79.4$\pm$1.1 & 50.6$\pm$0.3 \\
& ViT-B/16 & 83.4$\pm$1.2 & 53.9$\pm$1.2 \\
& ViT-L/14 & 87.6$\pm$1.2 & 59.0$\pm$0.6\\
\bottomrule
\end{tabular}
\end{sc}
\end{small}
\end{center}
\vskip -0.1in
\end{table}

As for Kinetics-600 dataset, as shown in \cref{table:zsl-kinetics-600}, we see that our method also performs favorably, especially compared to X-CLIP \cite{ni2022expanding} and TEXT4VIS \cite{wu2022transferring}.

\begin{table}[ht!]
\caption{Comparisons of zero-shot video action recognition performance of different algorithms on Kinetics-600.}
\label{table:zsl-kinetics-600}
\vskip 0.15in
\begin{center}
\begin{small}
\begin{sc}
\begin{tabular}{lccc}
\toprule
Method  & Top-1 Acc & Top-5 Acc \\
\midrule
\makecell{ER\\\cite{chen2021elaborative}} & 42.1$\pm$1.4 & 73.1$\pm$0.3 \\
\makecell{X-CLIP ViT-B/16\\\cite{ni2022expanding}} & 65.2$\pm$0.4 & 86.1$\pm$0.8\\
\makecell{Text4Vis ViT-L/14\\\cite{wu2022transferring}} & 68.9$\pm$1.0 & - \\
\midrule
\multirow{3}{*}{\system} \,\,\,ViT-B/32 & 69.5$\pm$0.6 & 91.7$\pm$0.1 \\
\quad\quad\quad\quad\quad\quad\quad ViT-B/16 & 73.0$\pm$0.8 & 93.2$\pm$0.1 \\
\quad\quad\quad\quad\quad\quad\quad ViT-L/14 & 81.1$\pm$0.8 & 96.3$\pm$0.3 \\
\bottomrule
\end{tabular}
\end{sc}
\end{small}
\end{center}
% \vskip -0.1in
\end{table}

\subsubsection{Results with Different Backbones}

\begin{table}[t!]
\caption{Comparing our method with various alternative methods using different backbone networks, including ViT-B/32, ViT-B/16 and ViT-L/14.  ``CLIP$^{*}$'' denotes the results of directly applying the image pretrained CLIP model. ``FINE-TUNE'' denotes the standard fine-tuning process using the same model as \system.}
\label{table:baseline-compare}
\vskip 0.15in
\begin{center}
\begin{small}
\begin{sc}
\begin{tabular}{lc|>{\columncolor{Gray}}ccc}
\toprule
% pretrain data
Net  & Dset&  CLIP$^*$ &  Fine-tune & \system \\
\midrule
\makecell{\multirow{3}{*}{B/32}} & UCF & 69.1 & 78.0\rise{8.9} & \,\,\textbf{79.5\rise{10.4}}  \\
& HMDB & 45.4 & 47.3\rise{1.9} & \textbf{49.9\rise{4.5}} \\
& K600 & 64.8 & 62.8\drop{2.0} & \textbf{69.5\rise{4.7}}\\
\midrule
\makecell{\multirow{3}{*}{B/16}} & UCF & 74.2 & 79.7\rise{5.5} & \textbf{83.5\rise{9.3}}\\
& HMDB & 47.6 & 49.2\rise{1.6} & \textbf{53.2\rise{5.6}}\\
& K600 & 68.1 & 65.9\drop{2.2} & \textbf{73.0\rise{4.9}}\\
\midrule
\makecell{\multirow{3}{*}{L/14}} & UCF &  80.5 & 85.0\rise{4.5} & \textbf{87.9\rise{7.4}}\\
& HMDB & 55.0 & 51.9\drop{3.1} & \textbf{58.3\rise{3.3}}\\
& K600 & 76.2 & 74.9\drop{1.3} & \textbf{81.1\rise{4.9}}\\
\bottomrule
\end{tabular}
\end{sc}
\end{small}
\end{center}
% \vskip -0.1in
% \vskip 0.1in
\end{table}
We compare \system  with the CLIP baseline and standard fine-tuning using various backbone networks. In particular, the results in the ``CLIP$^*$'' column of \cref{table:baseline-compare} show the performance of the pretrained CLIP model, which predicts each image frame of the video seperately. The ``FINE-TUNE'' column shows the results of a standard fine-tuning process using the same model as \system. We see from \cref{table:baseline-compare} that larger backbone networks (ViT-L/14 $>$ ViT-B/16 $>$ ViT-B/32) consistently improve zero-shot performance. For example, the results in the ``CLIP$^*$'' column show that the ViT-L/14 model outperforms the ViT-B/16 by 6.3\% (80.5\% vs 74.2\%), 7.4\% (55.0\% vs 47.6\%), and 8.1\% (76.2\% vs 68.1\%) on the UCF, HMDB, and Kinetics-600 datasets, respectively, while similar trends can be found in the last two columns of \cref{table:baseline-compare}, demonstrating that larger-scale vision-language pretrained models contain stronger zero-shot knowledge and are more robust when transferred from image to video tasks. However, as we can see from the results of ``FINE-TUNE'' and ``CLIP$^*$'' in \cref{table:baseline-compare}, the fine-tuning process does not always result in improved performance. For example, the fine-tuning process leads to worse zero-shot results on the HMDB dataset when using the ViT-L/14 backbone, and all zero-shot testing results on the Kinetics-600 show a performance drop after models are fine-tuned on Kinetics-400 under a standard paradigm.  
The results indicate that the knowledge stored in CLIP cannot always be transferred with conventional fine-tuning.

\begin{figure*}[ht]
    \centering
    \includegraphics[width=0.96\textwidth]{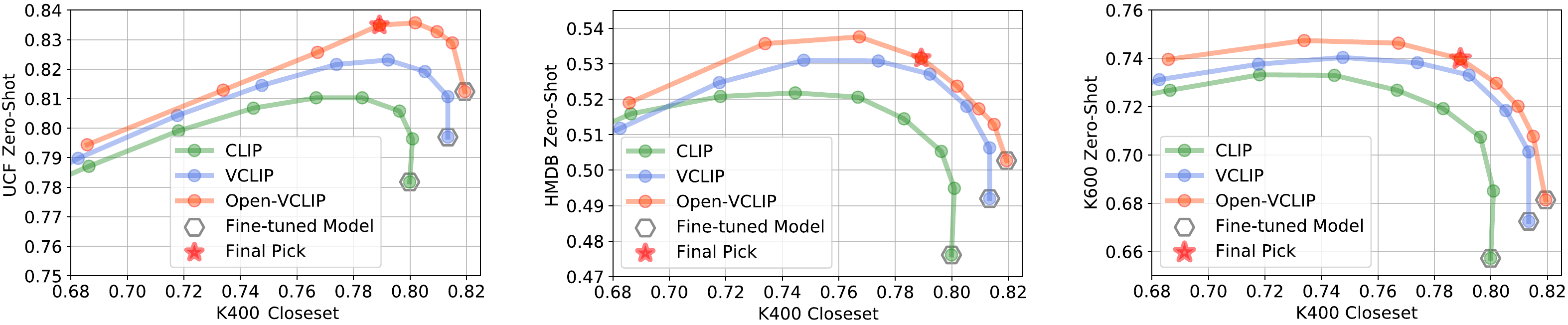}
         \caption{We evaluate the effectiveness of temporal modeling and weight interpolation with a VCLIP B/16 model. Points on each curve represent interpolation ratios of 0.8, 0.7, 0.6, 0.5, 0.4, 0.3, 0.2 and 0.0 from left to right, respectively. The red star marks correspond to our reported main results. We test on the full UCF, HMDB and the first split in~\cite{chen2021elaborative} on Kinetics-600. }
    \label{fig:discussion}
\end{figure*}

\begin{figure*}[ht!]
    \centering
    \includegraphics[width=0.96\textwidth]{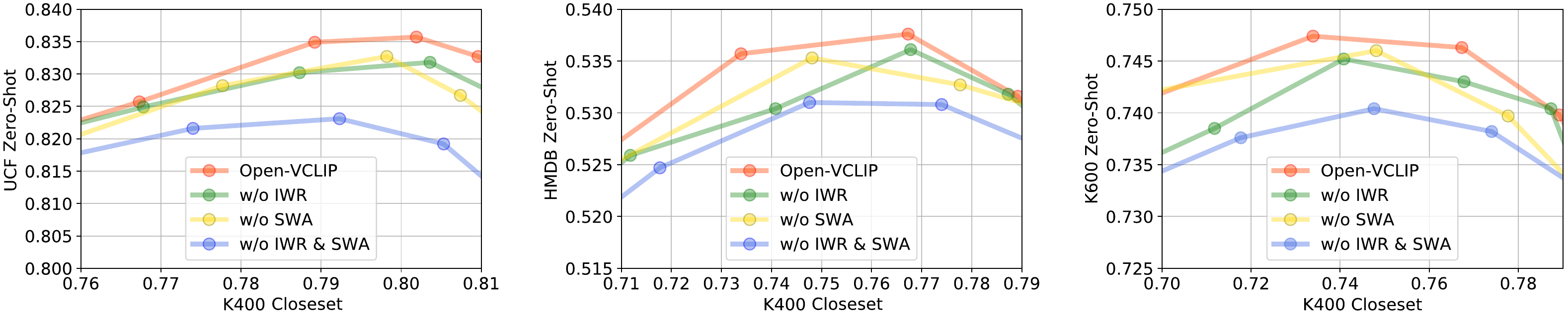}
     \caption{We evaluate the effectiveness of IWR and SWA with a VCLIP B/16 on the full UCF, HMDB and the first split in~\cite{chen2021elaborative} on Kinetics-600.} 
    \label{fig:ablation}
\end{figure*}
 
\begin{figure*}[ht!]
    \centering
    \includegraphics[width=0.95\textwidth]{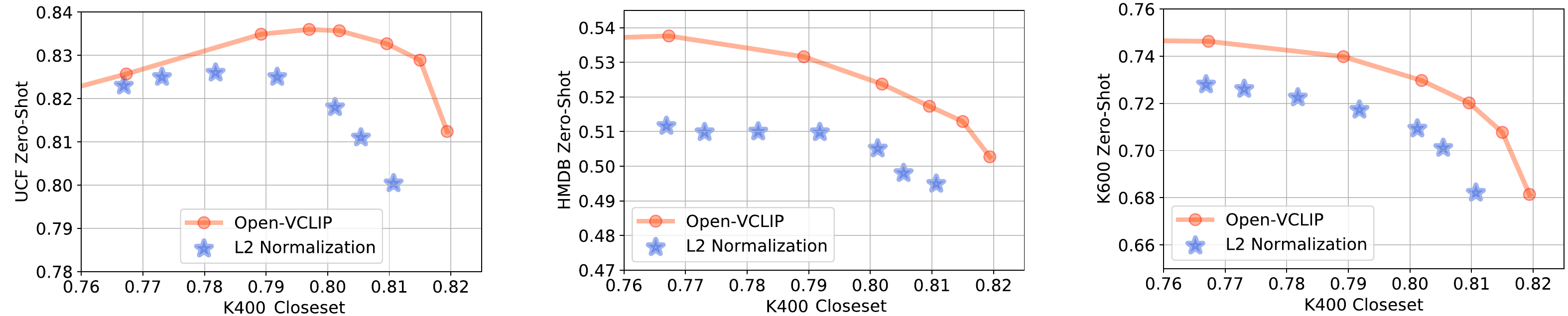}
     \caption{We compare Open-VCLIP with an $\ell_2$ regularization imposed on weigths on the full UCF, HMDB and the first split in~\cite{chen2021elaborative} on Kinetics-600.} 
    \label{fig:l2_regularization}
\end{figure*}

Compared to the baselines, our approach demonstrates significant and consistent improvements in zero-shot video action recognition across various datasets and models, as shown in the last column of Table~\ref{table:baseline-compare}. Taking experimental results on Kinetics-600 as an example, our approach not only avoids the performance drop seen in the ``FINE-TUNE'' method, but also significantly improves zero-shot recognition. This highlights the effectiveness of our method in effectively using annotated video datasets with limited labels for transfer learning and effectively preventing knowledge from forgetting, offering stable and decent results.

\subsection{Discussion and Ablation Studies}

\textbf{The effectiveness of temporal modeling and weight interpolation.}  We also investigate whether temporal modeling is needed for transferring CLIP to the video domain. To this end, we compare the performance of the CLIP, VCLIP, and \system, represented by green, blue and red curves in \cref{fig:discussion}, respectively. In particular, VCLIP expands the original spatial attention of CLIP to spatial-temporal attention without additional parameters as in \system. The curves illustrate weight interpolations with different mixing coefficients between CLIP and the fine-tuned model. Concretely, the $y$-axis displays the accuracy on the zero-shot video dataset, while the $x$-axis displays accuracy on the fine-tunning dataset, Kinetics-400.

Comparing the tails of the green and blue curves, we see that VCLIP achieves not only better close-set performance, but also better zero-shot performance on all of the three datasets. At the same time, when applying weight interpolation with different ratios, VCLIP achieves better trade-off than CLIP, as evidenced by the fact that the blue curve is always on top of the green curve in \cref{fig:discussion}. 
% This suggests the benefits of incorporating temporal modeling in zero-shot video action recognition.
This strongly suggests the significant advantages that come with incorporating temporal modeling in the context of zero-shot video action recognition.

Furthermore, the clear margins observed between the red curves and the other curves in \cref{fig:discussion} show that our proposed solution achieves the best trade-off compared to pure fine-tuning. With the same-level of close-set results, our approach always produces better zero-shot accuracy. In addition, the star marks on the red curves, which correspond to the reported main results in \cref{table:sota-compare,table:protocol3,table:zsl-kinetics-600}, are always better than the other baseline curves.

\textbf{IWR \vs SWA.} 
Our approach is composed of two weight interpolation modules: IWR which is used during training to regularize the fine-tuning process and SWA to improve generalization. We investigate their contributions to the final results in \cref{fig:ablation}. Overall, removing either IWR (green curve) or SWA (yellow curve) from the model leads to significant drops, \ie the red curve outperforms all other curves, suggesting IWR and SWA are complimentary to each other.
Furthermore, we see using only IWR or SWA is able to produce a good zero-shot performance improvement, compared to the results with no interpolation at all.

\begin{table*}[ht!]
\caption{Comparison of zero-shot video-to-text/text-to-video Retrieval performance for various algorithms. ``T2VRN'' denotes the recall@N of text-to-video retrieval. ``V2TRN'' denotes the recall@N of video-to-text retrieval. }
\label{table:vl-retrieval}
% \vskip 0.15in
\begin{center}
\begin{small}
\begin{sc}
\begin{tabular}{lccccccccc}
\toprule
Method  & K400 Tune & $\lambda$ & K400 & t2vR1 & t2vR5 & t2vR10 & v2tR1 & v2tR5 & v2tR10  \\
\midrule
Frozen~\cite{bain2021frozen} & $\times$ & - & - & 24.7 & 46.9 & 57.2 & - & - & - \\
CLIP~\cite{portillo2021straightforward} & $\times$ & - & - & 31.2 & 53.7 & 64.2 & 27.2 & 51.7 & 62.6 \\
CLIP (our implement) & $\times$ & - & 57.5 & 31.1 & 54.2 & 63.8 & 28.9 & 53.0 & 64.9 \\
\system & $\checkmark$ & 0.5 & 78.9	& 31.3&	54.3& 65.6 & 33.6 & 59.1 & 70.0 \\ 
\system & $\checkmark$ & 0.7 & 73.4 & 33.2 & 57.1 & 67.4 & 34.4 & 59.8 & 71.2\\
% \cmidrule(lr){3-4}\cmidrule(lr){5-6}
% & & EP1 & EP2 & EP1 & EP2 \\ 
% \midrule
% GA \cite{mishra2018generative} & C3D & 17.3$\pm$1.1 & - & 19.3$\pm$2.1 & -\\
% \midrule 
% \makecell{\multirow{3}{*}{\system}} & ViT-B/32 & 87.1$\pm$2.4 & 79.5 &  62.3$\pm$4.0 & 49.9  \\
%  & ViT-B/16 &  89.9$\pm$1.7 & 83.5 &  64.5$\pm$4.5 & 53.2 \\
%  & ViT-L/14 &  93.1$\pm$1.9 & 87.9 &  68.5$\pm$4.0 & 58.3 \\
\bottomrule
\end{tabular}
\end{sc}
\end{small}
\end{center}
% \vskip -0.1in
\end{table*}

\textbf{Weight regularization during fine-tuning.} IWR is conceptually similar to EWC~\cite{ewc}, which penalizes the changes of important parameters for solving previous tasks. However, EWC needs to assess the importance of parameters through historical data which is not feasible to our setting. Instead, we simply 
constrain the optimization by penalizing the weight changes using an $\ell_2$ norm. 
Concretely, we use the $\ell_2$ distance between the updated weights and the original weights as the regularization loss term during training, assigning various weights to the regularization loss term. As shown in \cref{fig:l2_regularization}, our method achieves the best trade-off. \system achieves higher close-set and zero-shot accuracy across all the datasets compared with simply applying an $\ell_2$ normalization, demonstrating the effectiveness of \system.

\textbf{Text-to-Video/Video-to-Text Retrieval Performance.} Assessing the model via text-to-video/video-to-text retrieval tasks offers insight into its generalizability within the video domain. We follow the paradigm of training models on Kinetics-400 dataset and testing them on MSR-VTT dataset~\cite{xu2016msr}, which is a large video description dataset. We firstly report the result of our own implementation using CLIP for retrieval as shown in the third row of~\cref{table:vl-retrieval}, which is similar to the results in~\cite{portillo2021straightforward}. This guarantees fair comparisons. Then, we evaluate \system on MSR-VTT.  We find that our text-to-video retrieval (given a text to retrieve the corresponding video) recall is comparable to that of raw CLIP but with a much higher close-set score on K400 (78.9\% vs 57.5\%). Further increasing the mixing coefficient $\lambda$ improves the text-to-video retrieval performance of \system, surpassing the CLIP baseline by 2\%. We also show that the video-to-text retrieval (given a video to retrieve the corresponding text) performance of \system is significantly higher than that of the CLIP baseline, demonstrating the effectiveness of our method in preserving the alignment capability when transferring to video domains.

% \textbf{Partially weights fine-tuning.}
% \hl{\textbf{Comparison with partial weights fine-tuning.} Fine-tuning a part of the network is an important and parameter-efficient method for adapting CLIP to a video model. 
% ST-Adapter~\cite{pan2022st} is a typical method that achieves good video classification performance by training the additional inserted adapter modules while keeping the original parameters of CLIP frozen. In practice, we find that ST-Adapter drops the text encoder of CLIP and introduces a linear classifier, which makes it not feasible for the model to do zero-shot testing. To address this issue, we keep the text encoder of CLIP when implementing ST-Adapter. Experiments in Table~\ref{table:compare_stadapter} demonstrate that ST-Adapter, which freezes the CLIP weights and fine-tunes only the added adapters, does not perform as well as our proposed method in terms of zero-shot action recognition. In particular, the zero-shot performance of ST-Adapter on K600 degrades significantly, demonstrating that partial weights fine-tuning does not solve the forgetting problem well.}

\textbf{Comparison with parameter-efficient fine-tuning.}
Fine-tuning only a part of the weights of a network is an effective approach for adapting the CLIP model to video in a parameter-efficient manner. A noteworthy method that employs this strategy is the ST-Adapter~\cite{pan2022st}, which accomplishes effective video classification by training added adapter modules, while keeping the original parameters of the CLIP model frozen. We compare ST-Adapter with our approach.
% to explore whether partial weights fine-tuning benefits the zero-shot video action recognition ability of CLIP.
More specifically, ST-Adapter forgoes the text encoder of CLIP, while introducing an additional linear classifier instead. This change prevents the model from being used in zero-shot testing. To circumvent this problem, we incorporate the text encoder of CLIP into our implementation of the ST-Adapter. The results presented in Table~\ref{table:compare_stadapter} illustrates that the ST-Adapter, while preserving the original CLIP weights and only fine-tuning the added adapters, fails to match the performance of our proposed method in zero-shot action recognition. In particular, we observe a marked degradation in the zero-shot performance of the ST-Adapter on the K600 dataset, suggesting that parameter-efficient fine-tuning does not effectively address the issue of catastrophic forgetting.

\begin{table}[h]
\caption{Comparison with parameter efficient fine-tuning method. }
\label{table:compare_stadapter}
\begin{center}
\begin{small}
\begin{sc}
\begin{tabular}{l|cccccccccc}
\toprule 
 Method & UCF & HMDB & K-600 \\
\midrule
CLIP & 74.2 & 47.6 & 68.1$\pm$1.0 \\
% \midrule 
Fine-tuned VCLIP & 79.7 & 49.2 & 65.9$\pm$1.0 \\
% \makecell{ST-Adapter\\\cite{pan2022st}} & 77.3 & 49.8 & 60.2$\pm$1.8 \\
ST-Adapter & 77.3 & 49.8 & 60.2$\pm$1.8 \\
\midrule
\system  & 83.5 & 53.2 & 73.0$\pm$0.8 \\
\bottomrule  
\end{tabular}
\end{sc}
\end{small}
\end{center}
% \vskip -0.3in
\end{table}

% \begin{table}[h]
% \caption{We compare our methods together. }
% \label{table:compare_stadapter}
% \begin{center}
% \begin{small}
% \begin{sc}
% \begin{tabular}{l|cccccccccc}
% \toprule 
%  Method & UCF & HMDB & K-600 \\
% \midrule
% \makecell{CLIP} & 74.2 & 47.6 & 68.1$\pm$1.0 \\
% % \midrule 
% \makecell{Fine-tuned VCLIP} & 79.7 & 49.2 & 65.9$\pm$1.0 \\
% % \makecell{ST-Adapter\\\cite{pan2022st}} & 77.3 & 49.8 & 60.2$\pm$1.8 \\
% \makecell{ST-Adapter} & 77.3 & 49.8 & 60.2$\pm$1.8 \\
% \midrule
% \makecell{\system}  & 83.5 & 53.2 & 73.0$\pm$0.8 \\
% \bottomrule  
% \end{tabular}
% \end{sc}
% \end{small}
% \end{center}
% % \vskip -0.3in
% \end{table}

\section{Conclusion}
We presented \system, an effective approach that enables CLIP to be transformed to an open-vocabulary video model. \system contains lightweight temporal modeling modules that equip CLIP with the ability to capture spatial and temporal relationships in videos. More importantly, \system is optimized with a carefully designed regularization strategy that strives for generalization to preserve the zero-shot abilities of CLIP. 
Extensive experiments are conducted and the results demonstrate that \system outperforms state-of-the-art methods with clear margins on zero-shot video action recognition and achieves the best trade-off between close-set and zero-shot video action recognition. One potential limitation is that adversaries could craft membership inference attacks to steal information from the model.

\section*{Acknowledgements}
This project was supported by NSFC under Grant No. 62032006 and No. 62102092. The authors would also like to thank Junke Wang for his help and suggestions.

\nocite{langley00}

\bibliography{example_paper}
\bibliographystyle{icml2023}

%%%%%%%%%%%%%%%%%%%%%%%%%%%%%%%%%%%%%%%%%%%%%%%%%%%%%%%%%%%%%%%%%%%%%%%%%%%%%%%
%%%%%%%%%%%%%%%%%%%%%%%%%%%%%%%%%%%%%%%%%%%%%%%%%%%%%%%%%%%%%%%%%%%%%%%%%%%%%%%
% APPENDIX
%%%%%%%%%%%%%%%%%%%%%%%%%%%%%%%%%%%%%%%%%%%%%%%%%%%%%%%%%%%%%%%%%%%%%%%%%%%%%%%
%%%%%%%%%%%%%%%%%%%%%%%%%%%%%%%%%%%%%%%%%%%%%%%%%%%%%%%%%%%%%%%%%%%%%%%%%%%%%%%
\newpage
\appendix
\onecolumn
\section{Proof}

\begin{lemma} Suppose the image CLIP model was trained on an image-text dataset $D_{\bar{A}} =\{\mathcal I_{\bar{A}},\mathcal T_{\bar{A}}\}$ with $N$ examples. Then, there exists a diverse video-text dataset $D_A=\{\mathcal V_A, \mathcal T_A\}$ containing $N$ examples where the video CLIP model with original CLIP parameters $\theta_A$ is optimal. 
\begin{proof}
We denote the large-scale visual-language pretraining dataset used to train CLIP as $D_{\bar{A}}=\{\mathcal I_{\bar{A}},\mathcal T_{\bar{A}}\}$. To bridge the gap between image and video domains, we extend images in $D_{\bar{A}}$ by repeating frames to create a set of static videos $\mathcal V_{A}=\{\mathcal V_i=[\mathcal I_{\bar{A}i}, \mathcal I_{\bar{A}i}, ..., \mathcal I_{\bar{A}i}]\}^N$ and let $\mathcal T_{A}=\mathcal T_{\bar{A}}$. Videos in $D_A=\{\mathcal V_A, \mathcal T_A\}$ are static  and do not provide any additional information for prediction. As a result, $\theta_A$ which offers the best results for $D_{\bar{A}}$ is also optimal on $D_A$.
\end{proof}
\end{lemma}

\end{document}